\documentclass[runningheads]{llncs}
\usepackage{graphicx}
\usepackage{comment}
\usepackage{amsmath,amssymb} 
\usepackage{color}
\usepackage{multirow}
\usepackage{epsfig}
\usepackage{booktabs}
\usepackage{hyperref}
\usepackage{enumitem}
\usepackage[multiple]{footmisc}
\usepackage[width=122mm,left=12mm,paperwidth=146mm,height=193mm,top=12mm,paperheight=217mm]{geometry}
\begin{document}
\pagestyle{headings}
\mainmatter

\title{RealMonoDepth: Self-Supervised Monocular Depth Estimation for General Scenes}

\titlerunning{RealMonoDepth}
%
\author{Mertalp Ocal and Armin Mustafa}
\authorrunning{Mertalp Ocal and Armin Mustafa}
%
\institute{Center for Vision, Speech and Signal Processing \\
University of Surrey, UK \\
\email{\{m.ocal,a.mustafa\}@surrey.ac.uk}}
\maketitle

\begin{abstract}

We present a generalised self-supervised learning approach for monocular estimation of the real depth across scenes with diverse depth ranges from 1--100s of meters. 
Existing supervised methods for monocular depth estimation require accurate depth measurements for training.  This limitation has led to the introduction of self-supervised methods that are trained on stereo image pairs with a fixed camera baseline to estimate disparity which is transformed to depth given known calibration.
Self-supervised approaches have demonstrated impressive results but
do not generalise to scenes with different depth ranges or camera baselines.   
In this paper, we introduce RealMonoDepth a self-supervised monocular depth estimation approach which learns to estimate the real scene depth for a diverse range of indoor and outdoor scenes. A novel loss function with respect to the true scene depth based on relative depth scaling and warping is proposed. 
This allows self-supervised training of a single network with multiple data sets for scenes with diverse depth ranges from both stereo pair and in the wild moving camera data sets.
A comprehensive performance evaluation across five benchmark data sets demonstrates that RealMonoDepth provides a single trained network which generalises depth estimation across indoor and outdoor scenes, consistently outperforming previous self-supervised approaches.\footnote{Code will be released upon publication.}\footnote{\url{https://youtu.be/6ot3hy3rGaA}}

\end{abstract}

\section{Introduction}

Humans comprehend 3D from a single viewpoint by leveraging the knowledge of context together with the shape and appearance priors. Similar to human visual perception, robust computer vision systems require the ability to perceive the environment in 3D. This fact has motivated research in monocular depth estimation. The problem to recover depth from a single image is ill-posed due to the projection ambiguity. Supervised methods have been proposed to estimate depth from a monocular image, demonstrating promising results, by training on a large amount of dense ground-truth depth data, however, this is expensive and impractical to acquire for real-world scenes \cite{dosovitskiy2015flownet,eigen2014depth,fu2018deep}. 
An alternative approach is to generate synthetic data by rendering from computer generated models \cite{planche2017depthsynth,chang2015shapenet,shilane2004princeton,xiang2014beyond} or 3D scans \cite{Choi2016,dai2017scannet,aanaes2016large}, but it is challenging to create data that represents the variety and detail of real-world appearance. 
Also transfer from training on synthetic data to real scenes remains a challenging open problem. Instead of regressing depth from raw pixels, self-supervised learning methods reformulate depth estimation as an image reconstruction problem by re-synthesising a target view from a single source view without ground-truth depth \cite{garg2016unsupervised,godard2017unsupervised}. Commonly these methods use stereo image pairs for training with a fixed camera baseline and require scenes with a fixed depth range. These methods learn to estimate inter-image disparity and then use camera calibration to estimate depth. 

\begin{figure}[t]
\centering
\includegraphics[width=100mm]{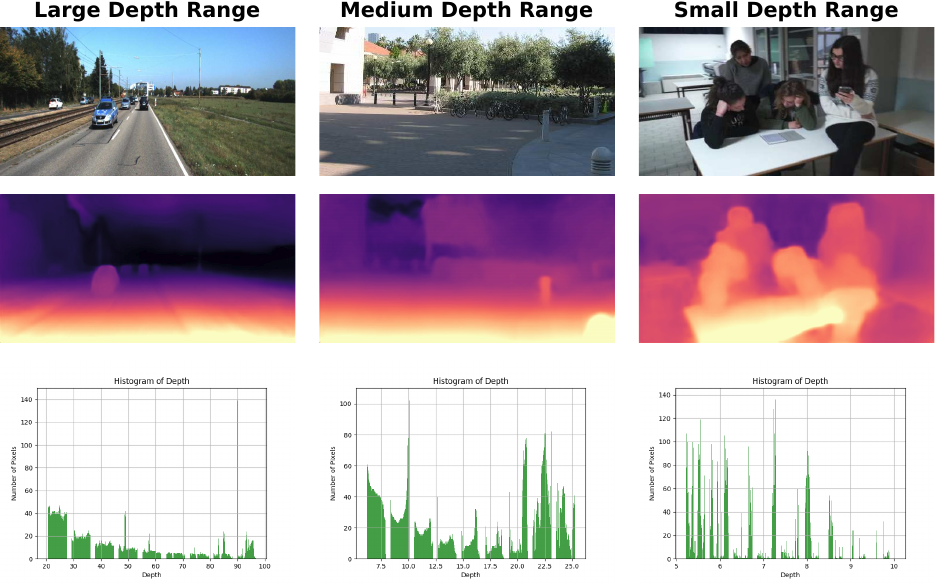}
\caption{\textbf{Top Row:} Scenes with different depth range: left, outdoor, range $\approx 500m$; middle, outdoor, range $\approx 25m$; and right, meeting room, range ($\approx 10m$). \textbf{Middle Row:} Depth from proposed method. \textbf{Bottom Row:} Depth histogram of ground-truth.}
\label{fig:example2}
\end{figure}

Learning monocular depth across diverse scenes is a challenging problem, due to large changes in depth range. Typically indoor scenes have a depth range of $<10m$, whereas outdoor scenes are commonly 100s of meters. 
Fig. \ref{fig:example2} shows some typical scenes with different depth ranges. Monocular depth estimation should be able to estimate depth across scenes with a wide variation in the depth range. It is infeasible to train a deep network that can regress depth values from raw pixels when the output space is incompatible. Existing self-supervised methods can be trained only on data sets with similar depth ranges \cite{godard2019digging,zhou2017unsupervised,mahjourian2018unsupervised,zhan2018unsupervised,pillai2019superdepth}, limiting the number of images that can be used for training. As a result, they demonstrate poor generalisation performance and can only perform specific tasks, such as depth estimation in outdoor driving scenes with a fixed stereo baseline. 
Recently, some supervised approaches \cite{li2019learning,li2018megadepth,chen2016single} addressed this issue by normalizing the ground-truth to have the same scale which allows learning relative depth values on moving camera data sets with a predefined depth range. However, these methods still depend on ground-truth depth values in order to estimate depth from an image.

\begin{figure}[t]
\centering
\includegraphics[width=110mm]{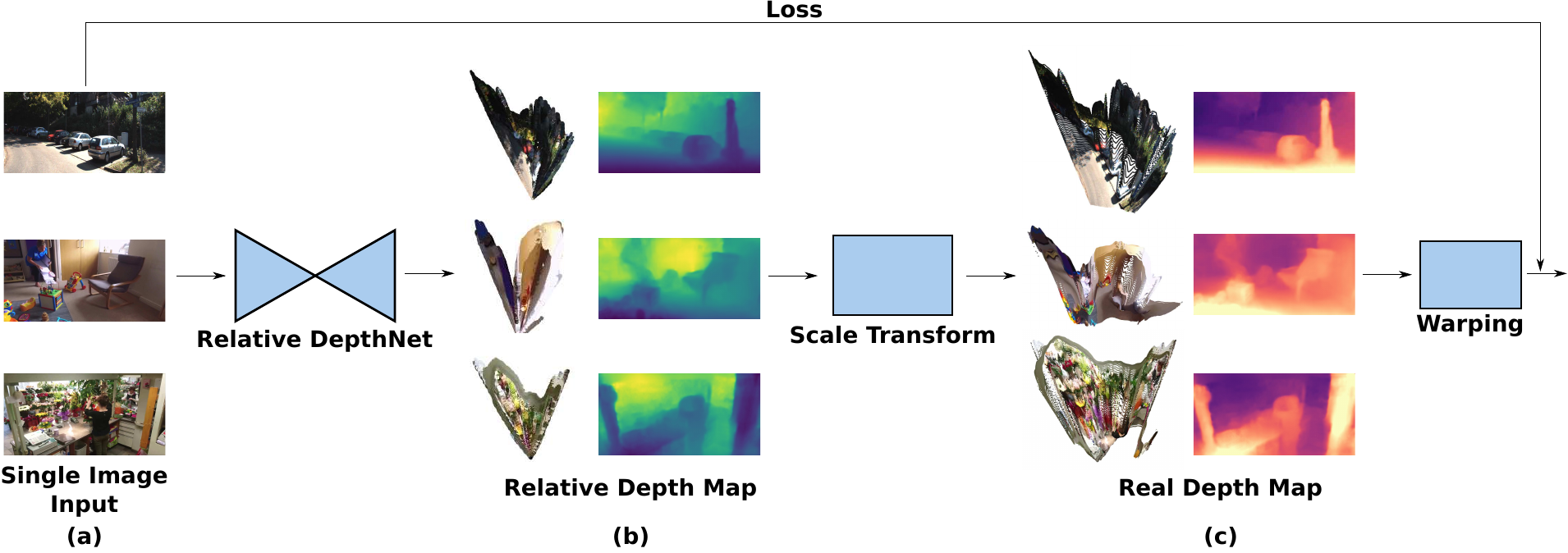}
\caption{Overview of RealMonoDepth approach for sample input images. For a given single view input image \textbf{(a)} the network predicts the \textbf{(b)} relative depth map, which  is scaled and warped to obtain \textbf{(c)} an estimate of the real depth to enable self-supervised training across scenes with diverse depth ranges.}
\label{fig:example0}
\end{figure}

In this paper, we propose a self-supervised method, RealMonoDepth, that allows a single network to estimate depth for indoor and outdoor scenes demonstrating improved accuracy over previous self-supervised monocular depth estimation approaches. The proposed network learns real depth from stereo-pair and moving camera data sets for scenes with a diverse depth range and without a fixed camera baseline. 
An overview of our approach for scenes with varying depth ranges (small, medium and large) is depicted in Fig. \ref{fig:example2}.
To enable self-supervised learning of depth estimation across multiple scene scales, we introduce a novel loss function to learn relative depth together with the real depth. The network learns the real depth by transferring relative depth through scaling and warping which is used to compute reconstruction loss for self-supervised training, as shown in Fig. \ref{fig:example0}. The contributions are of this work are:
\begin{itemize}[topsep=0pt,partopsep=0pt,itemsep=0pt,parsep=0pt]
\item A self-supervised monocular depth estimation method that is able to generalise learning across scenes with different depth range. 
\item A novel loss function over real depth for self-supervised learning of depth from a single image.
\item Evaluation on five benchmark data sets demonstrates generalisation across indoor and outdoor scenes with improved performance over previous work.
\end{itemize}

\section{Related Work}
Although it is not possible to perform metric reconstruction from a single image due to the projective scale ambiguity, recently proposed learning-based methods have demonstrated that a reliable estimate of the scene geometry can be generated by having prior knowledge of the scale of objects in the scene. This section reviews the supervised, unsupervised and self-supervised approaches that take a single RGB  image as input and estimate per-pixel depth as output.

\subsection{Supervised Depth Estimation}

Supervised single image dense depth prediction methods exploit depth values obtained from active sensors such as Kinect and LIDAR as ground-truth for training. Eigen et al. \cite{eigen2014depth} and Fu et al. \cite{fu2018deep} exploited regression loss to estimate depth for indoor and outdoor scenes respectively. Mayer et al. \cite{mayer2016large} demonstrated that training a fully convolutional network \cite{long2015fully} is an effective approach to learn disparity from stereo images. They also generated a large synthetic data set to train their network called DispNet, demonstrating improved performance on scenes with diverse depth ranges. However the network trained on synthetic data gives limited performance on real in the wild scenes.
Huang et al. \cite{huang2018deepmvs} proposed a deep CNN that leverages multi-view stereo images with known camera poses and calibration to generate patch plane sweep volumes with respect to a reference view. Matching is performed between these patches and produces the inverse disparity map for the reference view. Instead of generating plane sweep volumes, Yao et al. \cite{yao2018mvsnet} applied differential homography warping on 2D feature maps learned from multiple input views to produce a 3D cost volume. However, these methods suffer from the limited availability of multi-view data of real scenes with ground-truth depth, thereby relying heavily on synthetic data to train their network, which leads poor generalisation capability on complex real scenes. 
Hence these supervised methods require ground-truth depth or synthetic data for monocular depth estimation on real scenes.
\subsection{Unsupervised and Self-Supervised Depth Estimation}

Recently, unsupervised and self-supervised methods for depth estimation have gained attention eliminating the requirement of ground-truth depth data for real scenes. Unsupervised methods simultaneously estimate the depth and pose from a single RGB image and self-supervised methods exploit camera pose information estimated as a pre-process to obtain depth from a single monocular image. 

Zhou et al. \cite{zhou2017unsupervised} pioneered the work in unsupervised depth estimation by proposing separate deep CNN networks for pose estimation between unlabelled video sequences and for single view depth estimation.
Instead of training an additional pose estimator network, Wang et al. \cite{wang2018learning} implemented a differentiable version of Direct Visual Odometry Space (DVO) \cite{steinbrucker2011real} which is popularly used in current SLAM \cite{montemerlo2002fastslam,davison2003real} algorithms. Specifically, DVO solves for the pose by minimizing the warping loss of the reference frame from the source frame given the reference frame depth. Furthermore, they introduced a depth normalisation layer in order to address the scale ambiguity problem, which significantly outperformed \cite{zhou2017unsupervised}. 
Inspired by the relation between depth, pose and optical flow tasks in 3D scene geometry, Yin et al. \cite{yin2018geonet} jointly trained an unsupervised end-to-end deep network to predict pose and depth for non-rigid objects. An optical flow consistency check is imposed between backward and forward flow estimations for reliable estimation. 
However all these unsupervised methods give a limited performance on general scenes because of the ambiguity in the projection scale introduced by both unknown depth and pose.

Self-supervised methods exploit known camera pose information to resolve the depth ambiguity given estimates of disparity. Garg et al. \cite{garg2016unsupervised} introduced a self-supervised learning approach to train a deep network from stereo pairs by exploiting the epipolar relation between the cameras given a known calibration in order to generate inverse warp of the left view to reconstruct the right view. Although their results are impressive, they use non-differentiable Taylor series expansion to perform warping of disparity.
Godard et al. \cite{godard2017unsupervised,godard2019digging} imposed left-right consistency as a constraint for disparity regularisation and established a differentiable optimisation by leveraging spatial transformer networks \cite{jaderberg2015spatial} for bilinear sampling. Given a stereo pair as input for training, they estimate two disparity maps: left view disparity with respect to right view and right view disparity with respect to left view. Then, they reconstruct both views and also the disparity maps to achieve left-right consistency. Moreover, they also imposed edge-aware smoothness as another regulation along with left-right consistency and leveraged SSIM \cite{wang2004image} loss in addition to L1 reconstruction loss. Poggi et al. \cite{poggi2018learning} proposed an improvement to \cite{godard2017unsupervised} by leveraging three rectified views instead of stereo in order to establish additional disparity consistency. Inspired by recent successes of deep learning in single image super-resolution, Pillai et al. \cite{pillai2019superdepth} employ sub-pixel convolutional layers instead of resizing convolution layers. Contrary to previous methods that are limited to low-resolution operation, their method exploits high fidelity for better self-supervision. 
Existing self-supervised approaches estimate disparity and assume a fixed camera baseline during training. This limits the approaches to training for scenes with a similar depth range and does not allow generalisation across diverse indoor and outdoor scenes, or diverse data sets for training. In this paper, we introduce an approach to self-supervised learning using a loss function based on estimates of the true scene depth. This allows generalisation across both stereo pair and moving camera data sets for scenes with different depth ranges.

Both self-supervised and unsupervised methods suffer from the following limitations: requirement of fixed camera baseline; limited generalisation performance on scenes with varying depth ranges (indoor/outdoor); and self-supervised methods estimate disparity and only work for training on stereo image pairs with a fixed baseline which limits the training set. The proposed self-supervised depth estimation method addresses all of these limitations by generalising learning across scenes with different depth ranges and works for both stereo and moving camera data sets improving generalisation and accuracy of depth estimation over previous approaches.


\section{Method}

This paper introduces a self-supervised single image depth estimation approach that is able to generalise learning across scenes with diverse depth ranges. The method is trained on both stereo image pairs and moving camera data sets, giving state-of-the-art performance across five benchmark data sets. 
The method estimates depth from a single view image, an overview is shown in Fig. \ref{fig:example0}. The proposed network is trained on two views of a static scene. 
The Relative DepthNet network estimates relative depth maps from two views, inspired from \cite{godard2019digging} which estimates disparity maps between stereo pair. As a pre-process, camera calibration is estimated from sparse correspondences between two views using an existing visual SFM method COLMAP \cite{schoenberger2016sfm}. The sparse correspondences are used to estimate the median scene depth for scale transformation, which is applied on the relative depth maps to obtain real depth maps/true scene depth. The real depth maps and images are warped between views using the calibration to estimate the loss. This enables the network to be trained on moving camera and stereo data sets of real scenes with varying depth ranges and camera view baselines to generalise performance across a variety of indoor and outdoor scenes in the wild.

\begin{figure}[t]
\centering
\includegraphics[width=110mm,height=70mm]{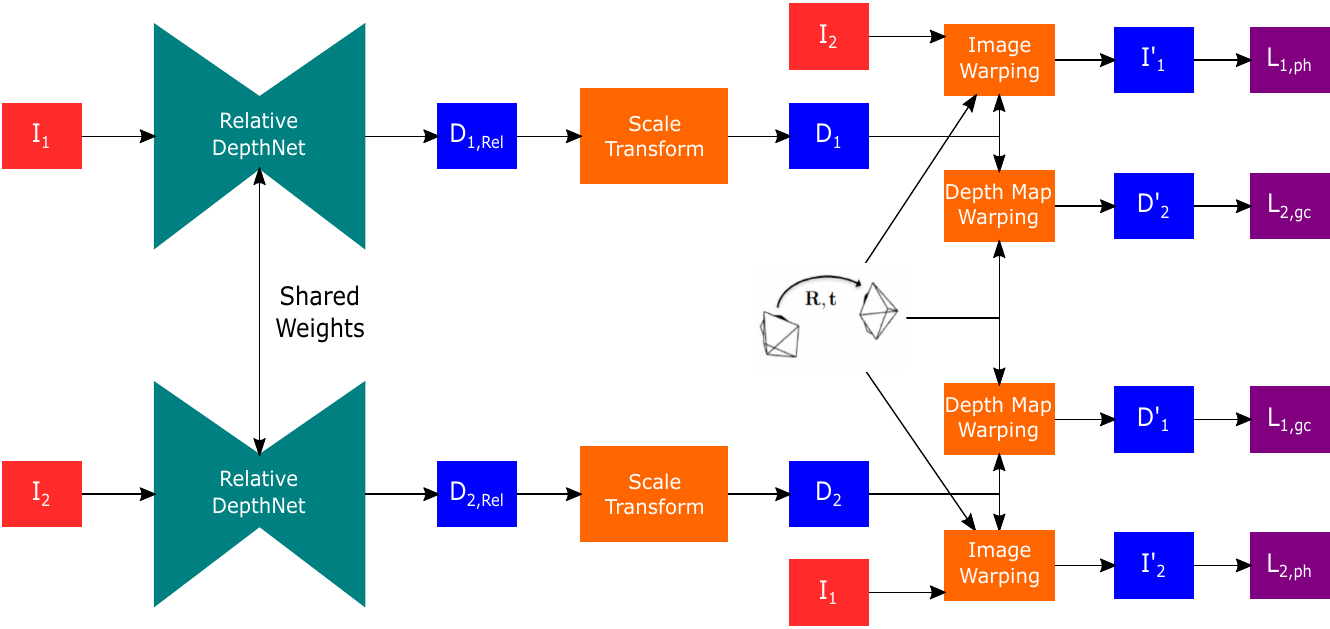}
\caption{Training framework for proposed self-supervised single image depth estimation}
\label{fig:example3}
\end{figure}

\subsection{Training Framework}
The training framework for the proposed approach is illustrated in Fig. \ref{fig:example3} for learning single view depth estimation from two viewpoint images. 
Given two images of a scene from different viewpoints (\textbf{$I_{1}$}, \textbf{$I_{2}$}), the depth network (Relative DepthNet) predicts the corresponding per-pixel relative depth maps (\textbf{$D_{1,Rel}$}, \textbf{$D_{2,Rel}$}) using shared weights. The relative depth maps are transformed to real depth (\textbf{$D_{1}$}, \textbf{$D_{2}$}) using the scale transform module. The self-supervised loss is then computed using the warped real depth estimates (\textbf{$D_{1}$}, \textbf{$D_{2}$}) and warped images (\textbf{$I_{2}$}, \textbf{$I_{1}$}). Using estimated calibration and camera poses, real depth values allow us to reconstruct the input images (\textbf{$I'_{1}$}, \textbf{$I'_{2}$}) and depth maps (\textbf{$D'_{1}$}, \textbf{$D'_{2}$}). This information is interpolated to compute photometric loss (\textbf{$L_{ph}$}) and geometric consistency (\textbf{$L_{gc}$}) loss, that supervises the depth network. SSIM and smoothness losses are introduced to regulate the depth estimation. The loss function for the proposed method is:

\begin{equation} \label{eqn:eq19}
{L_{final} = \sum_{s} \lambda_{ph}L^s_{ph}+\lambda_{gc}L^s_{gc}+\lambda_{ssim}L^s_{ssim}+\lambda^s_{s}L^s_{smooth}}
\end{equation}

\noindent 
where $s$ indexes over different image scales and $\lambda_{ph}$, $\lambda_{gc}$, $\lambda_{ssim}$ and $\lambda_{smooth}$ are the weighting terms.
Extension to training with multiple viewpoint images (>2) is straightforward (using loss $L_{i,ph},L_{i,gc}$ based on the estimated real scene depth $D_i$ for each input image $i$).
While existing self-supervised methods are only suited for training on fixed baseline cameras and scenes with similar depth range, we overcome this limitation by stabilizing the output space of the depth network, allowing training on indoor and outdoor scenes with diverse depth ranges. \\

\noindent
\textbf{Relative DepthNet Architecture:} 
\\
Based on the U-Net Architecture \cite{ronneberger2015u}, in order to effectively capture both local and global information, we use a multi-scale encoder-decoder network with skip connections, similar to \cite{godard2017unsupervised}. Inspired by \cite{guo2018learning,godard2019digging,kuznietsov2017semi}, we select ResNet50 \cite{he2016deep} for the encoder with initialised weights pre-trained on ImageNet \cite{russakovsky2015imagenet} and randomly initialise decoder weights. Unlike popular self-supervised/unsupervised single depth estimation approaches \cite{godard2017unsupervised,godard2019digging,mahjourian2018unsupervised,zhou2017unsupervised,pillai2019superdepth,li2018undeepvo}, our network predicts relative depth instead of inverse depth or disparity. 
We observe that applying sigmoid activation at the network output slows convergence for close and far depth values due to the vanishing gradient problem. Instead, we replace sigmoid with identity activation and handle negative values with exponential mapping, detailed in the Scale Transform section. Apart from multi-scale output layers of the decoder, we apply batch normalisation and ReLU nonlinearities in all layers. A detailed explanation with the full details of the network architecture required for implementation is presented in the supplementary material.\\ 

\noindent
\textbf{Scale Transform:} 
\\
To learn to estimate depth from images across diverse scenes with varying depth ranges, we normalise depth across the images and data sets using a non-linear scale transform and train the network to estimate relative depth. Given the relative depth map prediction as input, our scale transform module outputs the real depth map. This is formulated as:
\begin{equation} \label{eqn:st}
  D_{k} = {\mu_{k}}e^{(D_{k,Rel})} \text{ for } k= 1 , 2 
\end{equation}
where \textbf{$\mu_{k}$} is the median depth value for two images \textbf{$I_{1}$} and \textbf{$I_{2}$}, \textbf{$D_{1}$} and \textbf{$D_{2}$} are the real scene depth maps. Inspired by supervised methods \cite{eigen2014depth,fu2018deep}, we use exponential mapping in  equation \ref{eqn:st} in order to reduce the penalisation of the deep network from distant and ambiguous depth values.

During training, camera calibration is required to estimate the median depth value for the scale transform. For data sets with unknown calibration (e.g. Mannequin data set), off-the-shelf SFM method COLMAP \cite{schoenberger2016sfm} is used to solve for camera calibration and sparse correspondences between views. If the camera calibration is known (e.g. KITTI data set), we use the calibration to compute sparse correspondences between views. The sparse correspondences are triangulated in 3D exploiting camera pose to get a sparse reconstruction of the scenes. The sparse 3D points are projected on each view to obtain sparse depth maps for each viewpoint. Depth values are then sorted and the median depth value is estimated, as shown in Fig. \ref{fig:example2}. Median depth enables prediction of real depth maps which are used together with the input images to estimate the loss in Equation \ref{eqn:eq19}.\\

\noindent
\textbf{Loss Functions:} \\
\noindent
\textbf{Photometric consistency:} This loss enables estimation of depth and is inspired by self-supervised learning approaches that reformulate depth estimation as an image reconstruction problem \cite{garg2016unsupervised,godard2017unsupervised}. Here, the underlying intuition is that every viewpoint is a 2D projection of the same 3D scene, so one view can be reconstructed from another view, which implies knowledge about depth. 
With known calibration between views, depth is treated as an intermediate variable to perform novel-view synthesis with a deep network. Here,  $I'_{1}$ is the reconstructed reference view from \textbf{$I_{2}$} and $I'_{2}$ is the reconstructed reference view from \textbf{$I_{1}$}. Let \textbf{$K_{1}$}, \textbf{$K_{2}$} matrices represent the intrinsic parameters of \textbf{$I_{1}$} and \textbf{$I_{2}$} respectively, \textbf{$P_{1\,\to\,2}$}, \textbf{$P_{2\,\to\,1}$} denote relative camera pose matrices between views, \textbf{$D'_{2}(p'^{(2)})$}, \textbf{$D'_{1}(p'^{(1)})$} represent projected depth values for each pixel $p$. Then, the homogeneous pixel-wise projection relations ($D'_{proj}()$) between two views can be formulated as:
\begin{align} 
\label{eqn:eq11}
\begin{split}
{D'_{proj,2}(p'^{(2)})p'^{(2)} = K_{2}P_{1\,\to\,2}D_{1}(p^{(1)})K_{1}^{-1}p^{(1)}},
\\
{D'_{proj,1}(p'^{(1)})p'^{(1)} = K_{1}P_{2\,\to\,1}D_{2}(p^{(2)})K_{2}^{-1}p^{(2)}}
\end{split}
\end{align}
\noindent 
Since the projected pixel values are continuous, we use a spatial transformer network \cite{jaderberg2015spatial} to perform differentiable bilinear sampling in order to approximate $I'_{1}$ and $I'_{2}$ by interpolating \textbf{$I_{1}(p^{(1)})$} and \textbf{$I_{2}(p^{(2)})$} from neighboring corner pixels:

\begin{align*}
\begin{split}
{I'_{1}(p^{(1)}) = \textbf{$I_{2}(p^{(2)})$} = \sum_{(i,j)}w^{(2)}_{(i,j)}I_{2}(p'^{(2)}_{(i,j)})}
\text{ , }
{I'_{2}(p^{(2)}) = \textbf{$I_{1}(p^{(1)})$} = \sum_{(i,j)}w^{(1)}_{(i,j)}I_{1}(p'^{(1)}_{(i,j)})}
\end{split}
\end{align*}

\noindent 
Here, $(i,j)$ are the indices of the top left, top right, bottom left and bottom right pixels of the projected pixels and $w_{(i,j)}$ is the corresponding weight, which is inversely proportional to spatial distance. Also, due to occlusions between viewpoints, some pixels in the reference view will be projected outside of the image plane boundary in the source view. In order to prevent these unresolved regions from penalising the network, similar to Mahjourian et al. \cite{mahjourian2018unsupervised}, a validity binary masks is computed to exclude these pixels from the training loss \cite{godard2019digging}. Hence, the L1 image reconstruction loss for two views $k=1,2$ becomes:

\begin{align}  
\label{eqn:eq13}
\begin{split}
{L_{ph} = \frac{1}{N}\sum_p^N \sum_{k=1,2} L_{k,ph}(p)} \text{ where, }
{L_{k,ph}(p) = |I_k(p) - I'_{k}(p)|M_{k}(p)}
\end{split}
\end{align}

Also, SSIM \cite{wang2004image} loss is applied to 3x3 image patches in order to regulate the noisy artifacts caused by the L1 loss, defined as follows:
\begin{equation} \label{eqn:eq14}
L_{ssim} = \frac{1}{N}\sum_p^N[(1 - SSIM(I_1, I'_{1})(p)) + (1 - SSIM(I_2, I'_{2})(p))]
\end{equation}

\noindent \textbf{Smoothness:} 
This loss term ensures the depth maps are smooth, reducing the noise in the depth maps. Sobel gradients are used to calculate the loss instead of horizontal and vertical gradients in Wong et al. \cite{wong2019bilateral}. Sobel gradients allow depth to be smooth horizontally, vertically and diagonally. 
In most cases, regions that have higher reconstruction error are only visible in one view. Applying smoothness regularisation to these unresolvable regions induces a false penalty in the network training. In order to overcome this problem, an adaptive smoothness regulation weight is used for every pixel that varies in space, as in previous work \cite{wong2019bilateral}. Based on Equation \ref{eqn:eq13}, the adaptive weight for a pixel is computed as follows: ${\alpha_{k}(p) = exp(-\frac{c*L_{k,ph}(p)}{\sigma_{k}})}$, where index $k=1,2$ corresponds to the view and $c$ is the scale factor that determines the range of $\alpha$ and $\sigma$ is the global residual represented as: ${\sigma_{k} = \frac{1}{\frac{1}{N}\sum\limits_p^NL_{k,ph}(p)}}$.
$\alpha$ depends on global residual ($\sigma$) at each position and tends to be small when the residuals are high. The average of $\alpha$ approaches to 1 as the training converges. Based on adaptive weight, the smoothness regularisation objective for two views $k =1,2$ is:
\begin{equation}
    {L_{smooth} = {\frac{1}{N}\sum\limits_p^N \sum_{k=1,2} L_{k,smooth}(p)}}
\end{equation}
\begin{equation*}
    {L_{k,smooth}(p) = \alpha_{k}(p)(|\partial_xD_{k,rel}(p)|exp(-\partial_xI_{k}(p)) + |\partial_yD_{k,rel}(p)|exp(-\partial_yI_{k}(p))}
\end{equation*}

\noindent Here, the depth values are enforced to be locally smooth which is represented as the x and y Sobel gradients of the relative depth maps.

\noindent\textbf{Geometric consistency:} The per-view predicted depth maps may not be consistent with the same 3D geometry. This causes depth discontinuities and outliers on the surfaces and boundaries of the objects. In order to address this issue, we learn the real depth map (\textbf{$D_{1}$}, \textbf{$D_{2}$}) for each viewpoint simultaneously with consistency checks. Inspired by Bian et al. \cite{bian2019unsupervised}, we enforce geometric consistency symmetrically by sampling different viewpoints in the same training batch. Based on Equation \ref{eqn:eq11}, we interpolate projected depth maps $D'_{proj,1}$, $D'_{proj,1}$ with bilinear sampling in order to approximate $D'_{1}(p)$, $D'_{2}(p)$ which lie on the pixel grid. The geometric consistency loss function for $k=1,2$ is defined as follows:

\begin{equation} \label{eq:gc}
{L_{gc} = {\frac{1}{N}\sum\limits_p^N(L_{1,gc}(p) + L_{2,gc}(p))}} \text{ where }
{L_{k,gc}(p) = \frac{|D_{k}(p)-D'_{k}(p)|}{D_{k}(p)+D'_{k}(p)}} 
\end{equation}
 
\noindent Here, similar to \cite{bian2019unsupervised} we use a normalised symmetric loss function to achieve depth consistency between predicted depth maps for each viewpoint.

\begin{table}[ht]
\begin{center}
\caption{Data sets for monocular depth estimation}
\scalebox{0.8}{
\begin{tabular}{lcccccccc}
\toprule
data set & Indoor & Outdoor & Dynamic & Video & Depth & Diversity & Annotation & \# Images\\
\midrule
NYUDv2 \cite{silberman2012indoor} & \checkmark &  & \checkmark & \checkmark & Metric & Low & RGB-D & 407K\\
Make3D \cite{saxena2008make3d} &  & \checkmark &  &  & Metric & Low & Laser & 534\\
KITTI \cite{geiger2012we} & & \checkmark & \checkmark & \checkmark & Metric & Low & Stereo & 93K\\
DIW \cite{chen2016single} & \checkmark & \checkmark & \checkmark & & Ordinal Pair & High & User clicks & 496K\\
Cityscapes \cite{cordts2016cityscapes} & & \checkmark & \checkmark & \checkmark & Metric & Low & Stereo & 25K\\
Megadepth \cite{li2018megadepth} & & \checkmark & \checkmark & & No scale & Medium & SFM & 130K\\
MC \cite{li2019learning} & \checkmark & \checkmark & \checkmark & \checkmark & No scale & High & SFM & 115K\\
TUM \cite{sturm2012benchmark} & \checkmark &  & \checkmark & \checkmark & Metric & Low & RGB-D & 80K\\
\bottomrule
\end{tabular}
}
\label{Tab:tab2}
\end{center}

\end{table}%


\section{Experiments}

Qualitative and quantitative results are presented on five benchmark data sets against state-of-the-art supervised, unsupervised and self-supervised methods. We demonstrate that the proposed self-supervised loss function using real depth dramatically improves generalisation performance when trained on both moving camera (Mannequin Challenge (MC) \cite{li2019learning} mostly indoor) and stereo (KITTI \cite{geiger2012we} outdoor) data sets jointly. These data sets contain both indoor (1--10m) and outdoor (1--1000m) scenes with a wide variation in depth range. We test the same trained model on four benchmark data sets which the network has not seen during training: KITTI Eigen test split \cite{eigen2015predicting} (street scenes), Make3D \cite{saxena2008make3d} (outdoor buildings), NYUDv2 test split \cite{silberman2012indoor} (indoor) and dynamic subset of TUM-RGBD \cite{li2019learning} (humans in indoor environments). 
Key attributes of the data sets used in experiments are listed in Table \ref{Tab:tab2}.  
Additional qualitative results on a wide variety of in the wild scene images for the DIW \cite{chen2016single} data set are presented in the supplementary material, together with comparative performance evaluation on a diverse range of challenging in the wild videos in the supplementary video. 

\subsection{Implementation Details}

Our model is implemented in Tensorflow \cite{abadi2016tensorflow}, trained using the Adam \cite{kingma2014adam} optimiser for 25 epochs with an input/output resolution of $512 \times 256$ and a batch size of $12$. Each batch sample consists of different viewpoint images of the same scene. We set the number of viewpoint images as $2$ which leads to $12 \times 2 = 24$ images for each batch. Initial learning rate is set to $10^{-4}$ and it is decayed with a linear scheduler $LR(iteration) = initialLR*(1 - iteration/maxIteration)^{0.9}$. The weights for the loss terms are empirically determined as: $\lambda_{ph}=0.15, \lambda_{ssim}=0.85, \lambda^s_{smoothness}=0.01/s$ where $s$ is the downsampling factor for each scale, $\lambda_{gc}=0.1$ and scale factor for adaptive regularisation term is chosen as $5$, similar to \cite{wong2019bilateral}. For data augmentation, we perform horizontal flipping, random scaling, cropping with $50\%$ and apply random brightness, contrast, saturation on the rest $50\%$, with the same range of values as in \cite{godard2019digging}. Full details of network implementation are given in the supplementary material.

\subsection{Experimental Setup}

We train our model based on data split of Eigen et al. \cite{eigen2014depth} for KITTI and Li et al. \cite{li2019learning} for Mannequin Challenge (MC) data sets. We perform both individual and mixed training on these data sets with and without our proposed scale transform approach in order to evaluate the differences in generalisation capability and compare with previous methods. The underlying motivation for combining these data sets is threefold: 1) They are both large data sets suitable for self-supervised training, 2) Their sizes are comparable which makes them favourable to mix for joint training, 3) They represent different varieties of appearance: outdoor street scenes in KITTI and humans (mostly indoors) in MC. 

We select 23,488 stereo pairs of KITTI for training and the remaining 697 images are used as test set for evaluating single view depth estimation. Similar to train split of \cite{li2019learning}, we select 2463 scenes of MC for training. Since some of the videos on Youtube were deleted by the owners, we were not able to access all of the video URLs provided by \cite{li2019learning}, so our training set is slightly smaller. In order to ensure well balanced class distribution, we randomly sample 40 viewpoint images from each scene. Images are resampled for scenes that have a lower number of viewpoints. 

We quantitatively evaluate the single view depth estimation of our model following the error metrics of Eigen et al. \cite{eigen2014depth}: mean absolute relative error (Abs Rel), mean squared relative error (Sq Rel), root mean squared error (RMS), and root mean squared log10 error (RMS(log)). Following Zhou et al. \cite{zhou2017unsupervised}, we scale our relative single-view depth map predictions to match the median of ground-truth.

\subsection{Comparison with State-of-the-art} \label{comparisons}

This section presents quantitative results of the proposed scale transform method models trained on various data set combinations (KITTI, MC, MC+KITTI). Note: due to estimation of disparity rather than depth and assumption of a fixed camera baseline in previous self-supervised estimation methods \cite{godard2017unsupervised,gordon2019depth,poggi2018learning,garg2016unsupervised}, it is not possible to train for data sets with different scene scales.

\noindent\textbf{KITTI} We report the results for the test set of KITTI based on Eigen split \cite{eigen2014depth} in Table \ref{Tab:tab4}. Our model trained on MC+KITTI shows the best generalisation performance outperforming all state-of-the-art supervised methods which are not trained on KITTI and which are trained on large scale diverse data sets such as Chen et al. \cite{chen2019towards} and Li et al. \cite{li2018megadepth}. We also demonstrate that the training loss function based on real depth allows generalisation over data sets for indoor and outdoor scenes with different scales and allows the combination of data sets during training without degrading test performance. Our model trained on MC+KITTI outperforms other state-of-the-art self-supervised/unsupervised methods on KITTI even when they are trained on KITTI.

\begin{table}[ht]
\caption{Quantitative results on KITTI data set for different methods trained on various scenes (lower is better).}
\centering

\scalebox{0.7}{
\begin{tabular}{lllcccc}
\toprule
Method & Supervision & Training set &  Abs Rel & Sq Rel & RMS & RMS(log)\\
\midrule
Eigen \cite{eigen2014depth} & Depth & KITTI & 0.203 & 1.548 &  6.31  & 0.282\\
Liu \cite{liu2015deep} & Depth & KITTI &  0.202 & 1.614 &  6.52 & 0.275\\
Fu \cite{fu2018deep} & Depth & KITTI & \textbf{0.072} & \textbf{0.307} & \textbf{2.73} & \textbf{0.120}\\
Eigen \cite{eigen2014depth} & Depth & NYU & 0.521 & 5.016 &  10.37 & 0.510\\
Liu \cite{liu2015deep} & Depth & NYU &  0.540 & 5.059 &   10.10 & 0.526\\
Laina \cite{laina2016deeper} & Depth & NYU &  0.515 & 5.049 & 10.07 & 0.527\\
Liu \cite{liu2015deep} & Depth & Make3D & 0.362 & 3.465 &   8.70 & 0.447\\
Laina \cite{laina2016deeper} & Depth & Make3D &  0.339 & 3.136 &  8.68 & 0.422\\
Chen \cite{chen2016single} & Depth & DIW & 0.393 & 3.260 & 7.12 & 0.474\\
Li \cite{li2018megadepth} & Depth & Megadepth & 0.368 & 2.587
 &  6.68 & 0.414\\
 \midrule
 Garg \cite{garg2016unsupervised} & Pose & KITTI & 0.152 & 1.226 & 5.85 & 0.246\\
 Monodepth \cite{godard2017unsupervised} & Pose & KITTI & 0.148 & 1.334 &  5.93 & 0.247\\
 DDVO \cite{wang2018learning} & & KITTI & 0.151 & 1.257 & 5.58 & 0.228\\
 GeoNet \cite{yin2018geonet} & & KITTI & 0.149 & 1.060 & 5.57 & 0.226\\
 Struct2depth \cite{casser2019depth} & & KITTI & 0.141 & 1.026 & 5.29 & 0.215\\
Zhou \cite{zhou2017unsupervised} & & KITTI & 0.208 & 1.768 & 6.86 & 0.283\\
Zhou \cite{zhou2017unsupervised} & & CS & 0.267 & 2.686
 &  7.58 & 0.334\\
 3Net \cite{poggi2018learning} & Pose & KITTI & 0.129 & 0.996 & 5.28 & 0.223\\
 Monodepth2 \cite{godard2019digging}(640x192) & Pose & KITTI & 0.109 & 0.873 & \textbf{4.960} & 0.209\\
 Ours & Pose & KITTI & 0.109 & 0.928 & 4.99 & 0.199\\
 Ours & Pose & MC & 0.276 & 2.563 & 9.17 & 0.386\\
 \textbf{Ours} & Pose & MC+KITTI & \textbf{0.108} & \textbf{0.855} & 5.15 & \textbf{0.204}\\
 
\bottomrule
\end{tabular}
}
\label{Tab:tab4}
\end{table}%

\begin{figure}[ht]
\centering
\includegraphics[width=100mm]{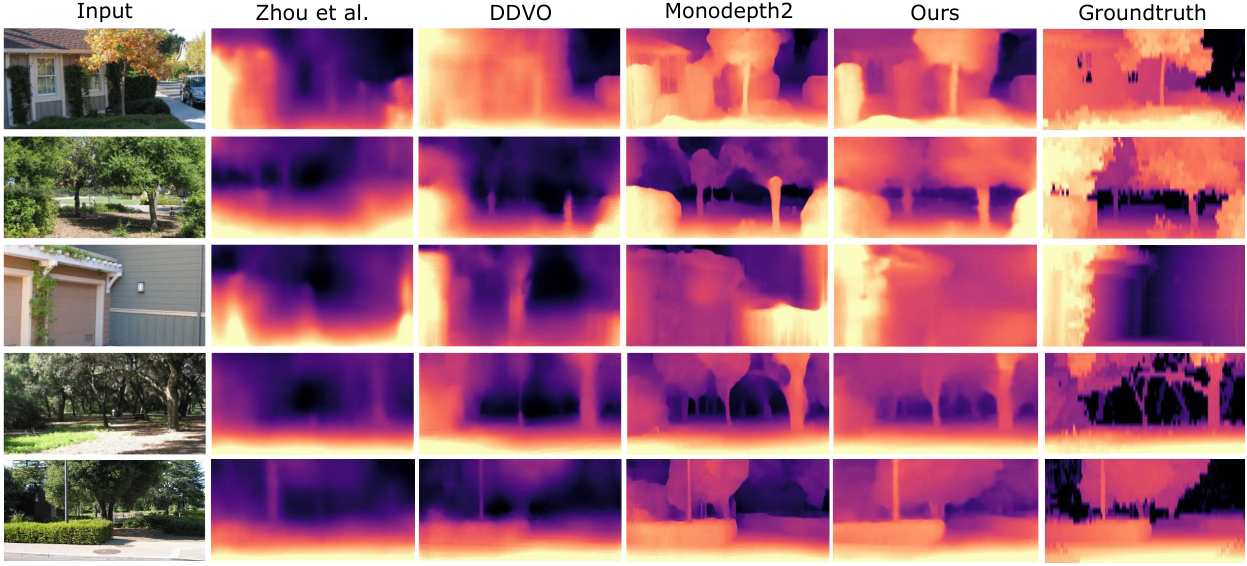}
\caption{Qualitative results on the Make3D data set.}
\label{fig:example5}
\end{figure}

\noindent\textbf{Make3D} Next, we evaluate on the Make3D data set following the procedure in \cite{godard2019digging}. In Table \ref{Tab:tab5}, our model trained on MC+KITTI demonstrates the best generalisation performance on an unseen data set compared to other methods which are \cite{liu2014discrete,liu2015deep} and are not \cite{eigen2014depth,chen2016single} trained on Maked3D. Qualitative comparisons of our method trained on MC+KITTI with training loss based on real depth demonstrates improved performance in Fig.\ref{fig:example5}.

\begin{table}[ht]
\caption{Quantitative results on Make3D and NYUDv2 data sets for different methods trained on various scenes (lower is better).}
\centering

\scalebox{0.7}{
\begin{tabular}{lllcc|lllcc}
\toprule
 \multicolumn{5}{c|}{Make3D} & \multicolumn{5}{c}{NYUDv2} \\
 Method & Supervision & Training set & Abs Rel & RMS & Method & Supervision & Training set & Abs Rel & RMS\\
\midrule
Xu \cite{xu2017multi} & Depth & Make3D & 0.184 & \textbf{4.38} & Xu \cite{xu2017multi} & Depth & NYU & \textbf{0.121} & 0.586\\
Li \cite{li2015depth} & Depth & Make3D & 0.278 &  7.19 & Li \cite{li2018monocular} & Depth & NYU & 0.139 & \textbf{0.505}\\
Laina \cite{laina2016deeper} & Depth & Make3D & \textbf{0.176} & 4.45 & Laina \cite{laina2016deeper} & Depth & NYU & 0.129 & 0.583\\
Liu \cite{liu2015deep} & Depth & Make3D & 0.314 & 8.60 & Liu \cite{liu2015deep} & Depth & NYU & 0.230 & 0.824\\
Liu \cite{liu2014discrete} & Depth & Make3D & 0.335 & 9.49 & Liu \cite{liu2014discrete} & Depth & NYU & 0.335 & 1.06\\
Laina \cite{laina2016deeper} & Depth & NYU & 0.669 & 7.31 & Eigen \cite{eigen2014depth} & Depth & NYU & 0.215 & 0.907\\
Liu \cite{liu2015deep} & Depth & NYU &  0.669 &  7.20 & Eigen \cite{eigen2015predicting} & Depth & NYU & 0.158 & 0.641\\
Eigen \cite{eigen2014depth} & Depth & NYU & 0.505 &  6.89 & Roy \cite{roy2016monocular} & Depth & NYU & 0.187 & 0.744\\
Chen \cite{chen2016single} & Depth & DIW & 0.550 &  7.25 & Wang \cite{wang2015towards} & Depth & NYU & 0.220 & 0.745\\
Li \cite{li2018megadepth} & Depth & Megadepth & 0.402 &  6.23 & Jafari \cite{jafari2017analyzing} & Depth & NYU & 0.157 & 0.673\\
\midrule
Monodepth \cite{godard2017unsupervised} & Pose & KITTI & 0.525 & 9.88 & Monodepth2 \cite{godard2019digging} & Pose & KITTI & 0.342 & 1.183\\
Monodepth2 \cite{godard2019digging} & Pose & KITTI & 0.322 & 7.42 & Ours & Pose & KITTI & 0.300 & 1.005\\
Zhou \cite{zhou2017unsupervised} & & KITTI & 0.651 &   8.39 & Ours & Pose & MC & 0.201 & 0.718\\
DDVO \cite{wang2018learning} & & KITTI & 0.387 & 8.09 & \textbf{Ours} & Pose & MC + KITTI & \textbf{0.193} & \textbf{0.686}\\
Ours & Pose & KITTI & 0.295 & 7.10 &  &  &  &  & \\
Ours & Pose & MC & 0.346 & 7.70 & & & & & \\
\textbf{Ours} & Pose & MC+KITTI & \textbf{0.289} & \textbf{6.92} & & & & & \\
\bottomrule
\end{tabular}
}
\label{Tab:tab5}
\end{table}%

\begin{figure}[h!]
\centering
\includegraphics[width=90mm]{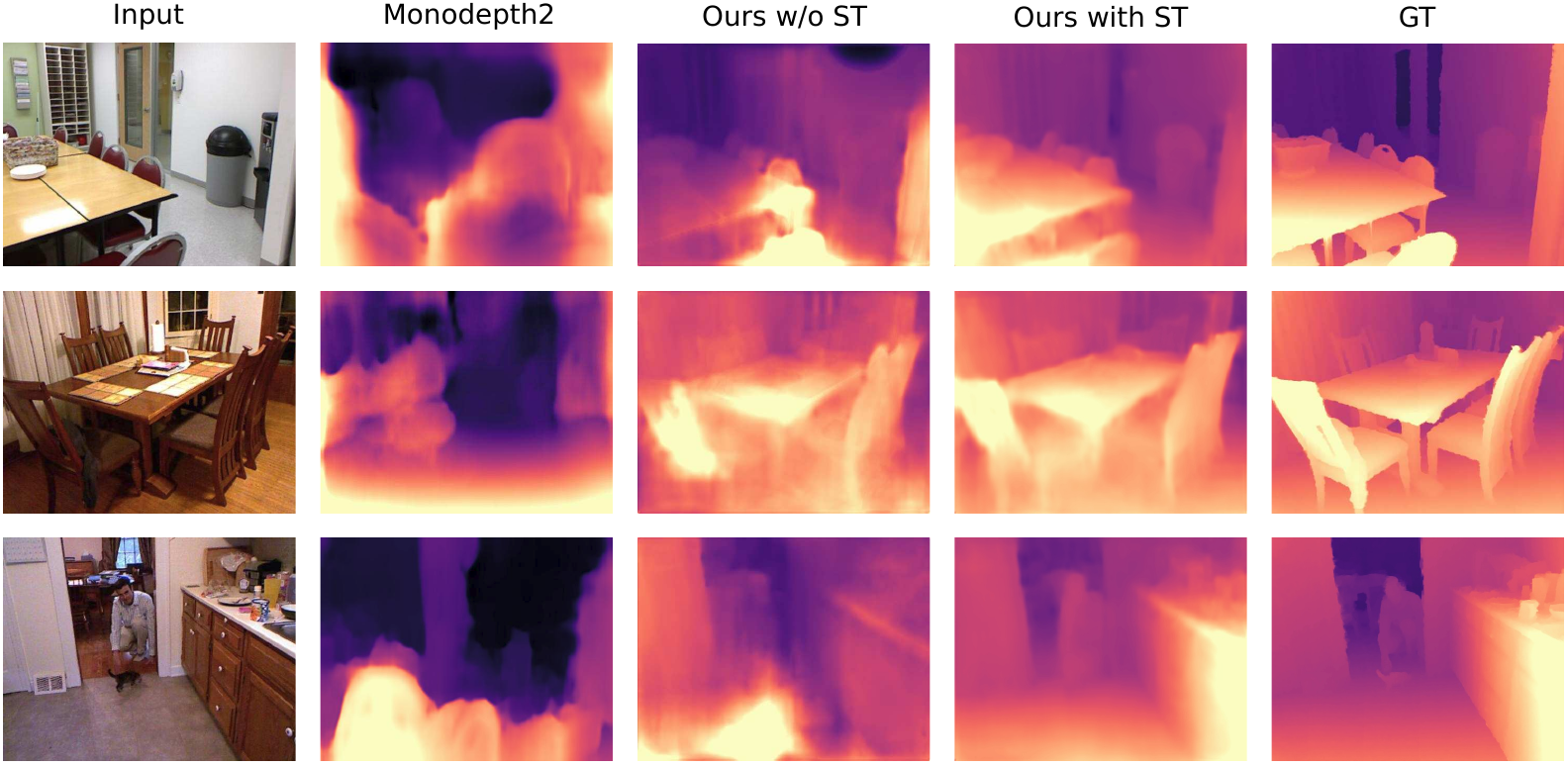}
\caption{Qualitative results on NYU data set.}
\label{fig:example6}
\end{figure}

\begin{figure}[h!]
\centering
\includegraphics[width=90mm]{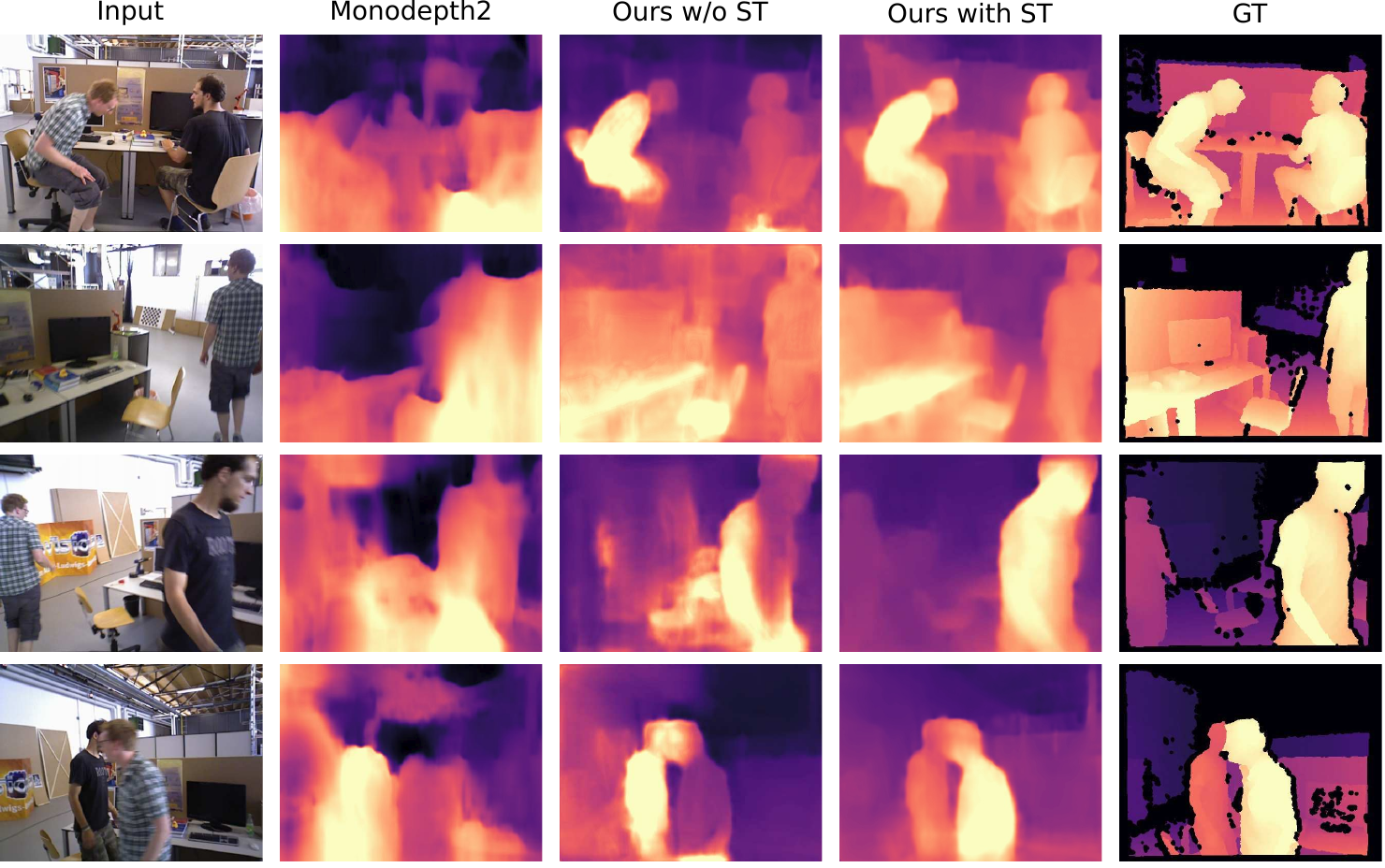}
\caption{Qualitative results on TUM data set.}
\label{fig:example7}
\end{figure}

\noindent\textbf{NYUDv2} Here, we show that our proposed method also generalises well to indoor scenes. We evaluate performance against state-of-the-art self-supervised monocular method Monodepth2\cite{godard2019digging} on the NYUDv2 test split. In Table \ref{Tab:tab5}, our model outperforms Monodepth2 with a significant margin and achieves competitive accuracy against supervised methods that are trained on a different split of the same data set. We also provide qualitative comparisons of our model trained on MC+KITTI in Fig. \ref{fig:example6} demonstrating improved performance.\\
\noindent\textbf{TUM} Finally, we present quantitative and qualitative results tested on the dynamic subset of the TUM-RGBD data set in Table \ref{Tab:tab7} and Fig. \ref{fig:example7} respectively. Our model ranks second-best compared to supervised methods and best for self-supervised methods. In order to make a fair comparison with \cite{li2019learning}, we only include their result trained on a single image without any additional prior knowledge except the depth ground-truth.

\begin{table}[h!]
\caption{Quantitative results on TUM Dynamic Objects RGBD data set for different methods trained on various scenes. Lower is better.}
\centering
\scalebox{0.7}{
\begin{tabular}{lllcc}
\toprule
 Method & Supervision & Training set & Abs Rel & RMS\\
\midrule
Xu \cite{ricci2018monocular} & Depth & NYU & 0.274 & 1.085\\
Laina \cite{laina2016deeper} & Depth & NYU & 0.223 & 0.947\\
Chen \cite{chen2016single} & Depth & NYU+DIW & 0.262 & 1.004\\
DeMoN \cite{ummenhofer2017demon} & Depth & TUM RGBD+MVS & 0.220 & 0.866\\
Fu \cite{fu2018deep} & Depth & NYU & \textbf{0.194} & 0.925\\
Li \cite{li2019learning}(single image) & Depth & MC & 0.204 & \textbf{0.840}\\
\midrule
Monodepth2 \cite{godard2019digging} & Pose & KITTI & 0.427 & 1.616\\
Ours & Pose & KITTI & 0.388 & 1.533\\
Ours & Pose & MC & 0.207 & 0.973\\
\textbf{Ours} & Pose & MC+KITTI & \textbf{0.201} & \textbf{1.025}\\
\bottomrule
\end{tabular}
}
\label{Tab:tab7}
\end{table}%

\subsection{Ablation Study}

\begin{figure}[h!]
\centering
\includegraphics[width=110mm]{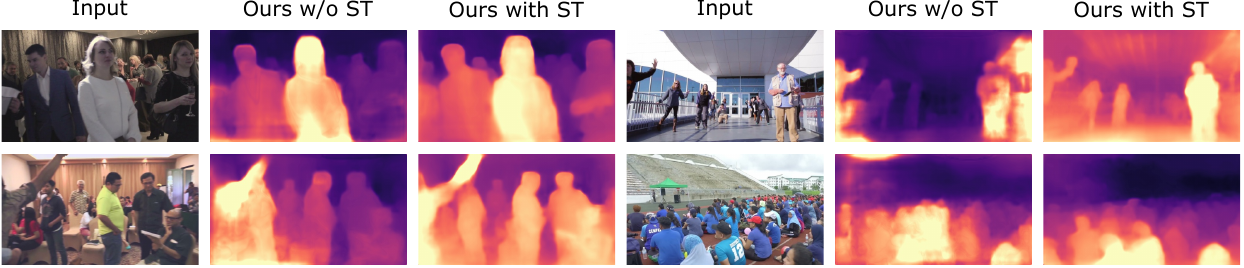}
\caption{Qualitative results on Mannequin Challenge data set.}
\label{fig:example8}
\end{figure}

Here, we evaluate the effect on generalisation performance across scenes with different depth ranges for our self-supervised training using a loss function based on real depth estimates. For models trained without our proposed method, we omit the scene median depth term, so the scale transform module is modified as $D = exp(D_{rel})$ where $D_{rel}$ is the estimated relative depth of our Relative DepthNet network and $D$ is the real depth map which is used for warping the other viewpoint image in order to compute the loss for training. We train our network with four different configurations regarding training data sets and usage of proposed scale transform (ST) during training: 1) MC w/o ST, 2) MC+KITTI w/o ST, 3) MC w/ ST and 4) MC+KITTI w/ ST. Then, we test each of these four models on four benchmark data sets similar to section 4.3. We report numerical ablation results in Table \ref{Tab:tab3} and show qualitative comparisons between models trained on MC+KITTI with and without scale transform in Fig. \ref{fig:example6}, Fig. \ref{fig:example7} and Fig. \ref{fig:example8}. Both our models trained on MC+KITTI with and without scale transform achieve similar numerical results on test split of KITTI. This is reasonable since the KITTI data set is collected with stereo cameras that have a fixed baseline between them. On the other hand, the MC data set consists of diverse scenes with no fixed baseline between viewpoint images in each scene. For other test sets, models trained with the proposed loss function significantly outperform the models that are trained without the scale transform. Moreover, our proposed method also allows training on a combination of different data sets to generalise across scene depth ranges for indoor and outdoor scenes.

\begin{table}[h!]
\caption{ Results on two different test sets with and without proposed scale transform method. Lower is better for all error measures.}
\centering
\scalebox{0.7}{
\begin{tabular}{llcccc|lllllc}
\toprule
\multirow{3}{*}{Test set} & \multirow{3}{*}{Error Measure} & \multicolumn{4}{c|}{Training Set}                        & \multicolumn{1}{c}{\multirow{3}{*}{Test set}} & \multicolumn{1}{c}{\multirow{3}{*}{Error Measure}} & \multicolumn{4}{c}{Training Set}                                                                                         \\ \cline{3-6} \cline{9-12} 
                          &                                & \multicolumn{2}{c}{MC} & \multicolumn{2}{c|}{MC + KITTI} & \multicolumn{1}{c}{}                          & \multicolumn{1}{c}{}                               & \multicolumn{2}{c}{MC}                                          & \multicolumn{2}{c}{MC + KITTI}                         \\
                          &                                & w/o ST     & ST        & w/o ST     & ST                 & \multicolumn{1}{c}{}                          & \multicolumn{1}{c}{}                               & \multicolumn{1}{c}{w/o ST} & \multicolumn{1}{c}{ST}             & \multicolumn{1}{c}{w/o ST} & ST                        \\ \hline
\multirow{2}{*}{KITTI}    & Abs Rel                        & 0.426      & 0.276     & 0.116      & \textbf{0.108}     & \multirow{2}{*}{NYU}                          & Abs Rel                                            & 0.282                      & 0.201                              & 0.333                      & \textbf{0.193}            \\
                          & Sq Rel                         & 4.933      & 2.563     & 0.864      & \textbf{0.855}     &                                               & RMS                                                & 0.921                      & 0.718                              & 1.347                      & \textbf{0.686}            \\ \hline
\multirow{2}{*}{Make3D}   & Abs Rel                        & 0.424      & 0.346     & 0.347      & \textbf{0.289}     & \multirow{2}{*}{TUM Dynamic}                  & Abs Rel                                            & 0.252                      & 0.207                              & 0.273                      & \textbf{0.201}            \\
                          & RMS                            & 8.771      & 7.70      & 7.64       & \textbf{6.92}      &                                               & RMS                                                & 1.050                      & \multicolumn{1}{c}{\textbf{0.973}} & 1.133                      & \multicolumn{1}{l}{1.025} \\ \bottomrule
\end{tabular}
}
\label{Tab:tab3}
\end{table}%
\section{Conclusion and Future Work}

We present a generalised self-supervised monocular depth estimation method (RealMonoDepth) that overcomes the limitation of existing self-supervised methods (\cite{gordon2019depth,zhou2017unsupervised,godard2017unsupervised,yin2018geonet,garg2016unsupervised}) that are limited to scenes with fixed scale and depth range. These methods cannot be trained on moving camera data sets due to the assumption of a fixed baseline and are unable to generalise to unseen data sets with different depth ranges.
RealMonoDepth addresses all of these limitations by allowing simultaneous training on a combination of indoor and outdoor scenes with varying depth ranges. This leads to significantly improved generalisation performance across indoor and outdoor scenes and scenes which are unseen during training,  and removes the dependance on a fixed camera baseline. 
The proposed method allows mixing stereo and moving camera data sets (MC + KITTI) improving on  state-of-the-art performance in single view depth estimation across five benchmark data sets including data sets with varying depth range.

Success of deep networks depends on the use of large data sets. The proposed self-supervised training from sequences captured from a single camera allows us to train the network on diverse uncontrolled in the wild data sets, such as the Mannequin Challenge (MC) data set used in this work. This opens the door to further generalisation through training across even larger and more diverse scenes.
A limitation of the proposed method is that it works with static scenes during training as with all the other single-image depth estimation methods. At test time we estimate depth from a single image and are therefore able to handle dynamic scenes with a wide variety of scale (see supplementary video). An interesting potential future work might be to extend the training of our method for dynamic scenes to increase the diversity of the data.
%
%
%
%
\bibliographystyle{splncs04}
\bibliography{eccv2020submission}

\title{Supplementary Material}
\titlerunning{}
\authorrunning{}
\author{}
\institute{}
\maketitle
\section{Network Implementation Details}

We use the standard pretrained resnet50 encoder, $resnet\_v1\_50.ckpt$, officially provided by Tensorflow [2]. The decoder weights are initialised randomly and the details of our architecture are shown in Table \ref{Tab:tab30}. Our decoder uses skip connections from the encoder [36] $(econv4, econv3,econv2, econv1)$, $econv5$ is the final encoder output and estimates multi-resolution depth maps ($depth1$, $depth2$, $depth3$, $depth4$) in order to exploit both local and global information to resolve higher resolution details. Code will be released upon publication.

\begin{table}[ht]
\caption{\textbf{Decoder architecture of Relative DepthNet.} $Upsample()$ represents $2\times$ nearest-neighbor resizing operation and $+$ denotes channel-wise concatenation.}
\centering
\scalebox{0.8}{
\begin{tabular}{c|c|c|c|c|c|c}
\toprule
Layer   & Output Size    & Kernel Size & Stride & Input                               & BatchNorm & Activation \\
\hline
upconv5 & $16 \times~ 32 \times$ 256  & 3           & 1      & Upsample(econv5)                    & Yes       & ReLU       \\
iconv5  & $16 \times~ 32 \times 256$  & 3           & 1      & upconv5 + econv4                    & Yes       & ReLU       \\
\hline
upconv4 & $32 \times~ 64 \times 128$  & 3           & 1      & Upsample(iconv5)                    & Yes       & ReLU       \\
iconv4  & $32 \times~ 64 \times 128$  & 3           & 1      & Upsample(upconv4) + econv3          & Yes       & ReLU       \\
depth4  & $32 \times~ 64 \times 1$    & 3           & 1      & iconv4                              & No        & Identity   \\
\hline
upconv3 & $64 \times~ 128 \times 64$  & 3           & 1      & Upsample(iconv4)                    & Yes       & ReLU       \\
iconv3  & $64 \times~ 128 \times 64$  & 3           & 1      & upconv3 + econv2 + Upsample(depth4) & Yes       & ReLU       \\
depth3  & $64 \times~ 128 \times 1$   & 3           & 1      & iconv3                              & No        & Identity   \\
\hline
upconv2 & $128 \times~ 256 \times 32$ & 3           & 1      & Upsample(iconv3)                    & Yes       & ReLU       \\
iconv2  & $128 \times~ 256 \times 32$ & 3           & 1      & upconv2 + econv1 + Upsample(depth3) & Yes       & ReLU       \\
depth2  & $128 \times~ 256 \times 1$  & 3           & 1      & iconv2                              & No        & Identity   \\
\hline
upconv1 & $256 \times~ 512 \times 16$ & 3           & 1      & Upsample(iconv2)                    & Yes       & ReLU       \\
iconv1  & $256 \times~ 512 \times 16$ & 3           & 1      & upconv1 + Upsample(depth2)          & Yes       & ReLU       \\
depth1  & $256 \times~ 512 \times 1$  & 3           & 1      & iconv1                              & No        & Identity \\
\bottomrule
\end{tabular}
}
\label{Tab:tab30}
\end{table}

\section{Additional Results}

\begin{figure}[!ht]
\centering
\includegraphics[width=122mm]{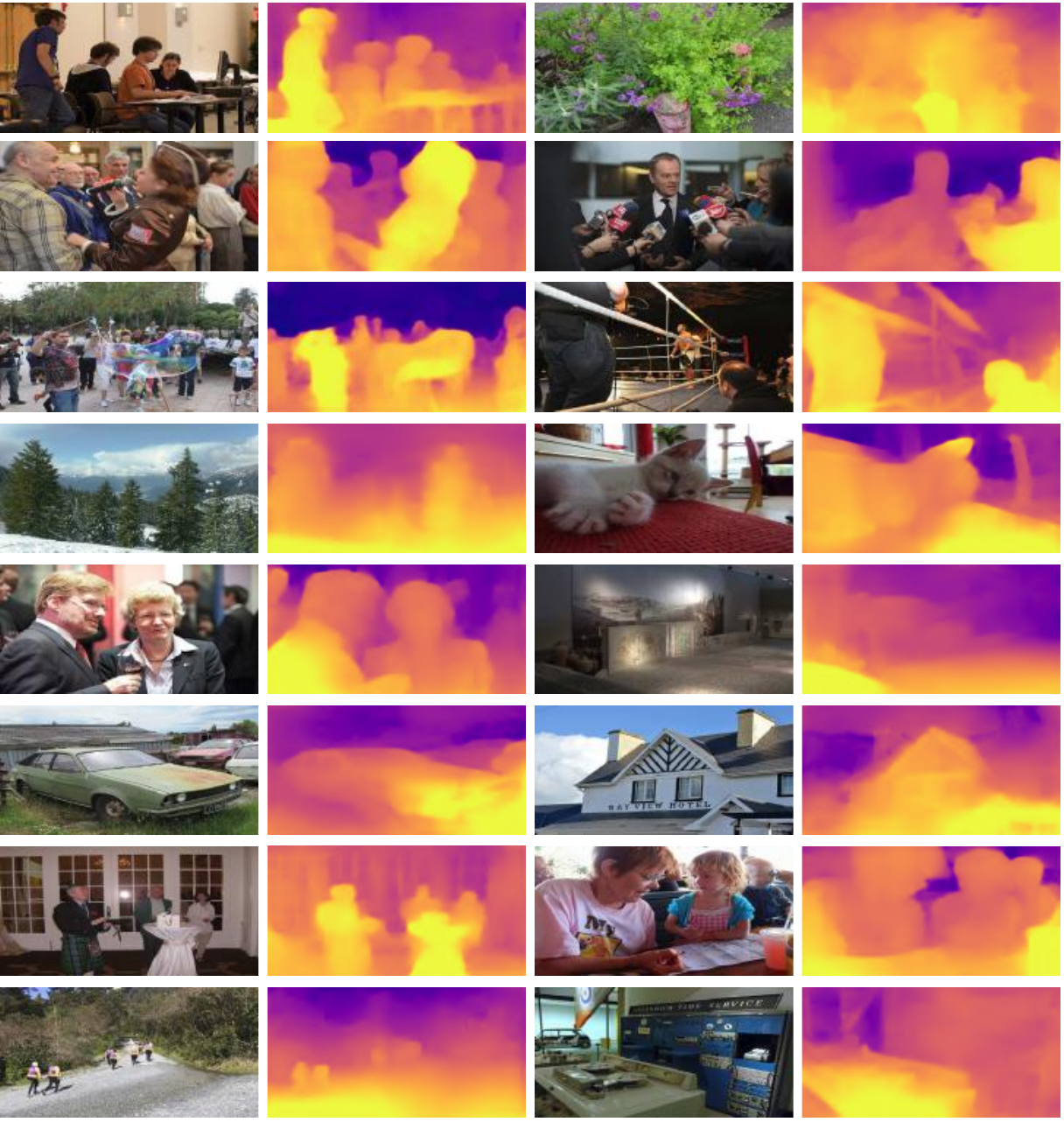}
\caption{Qualitative results on the DIW test set. \emph{(cont.)}}
\label{fig:example30}
\end{figure}

\begin{figure}[!ht]
\centering
\includegraphics[width=122mm]{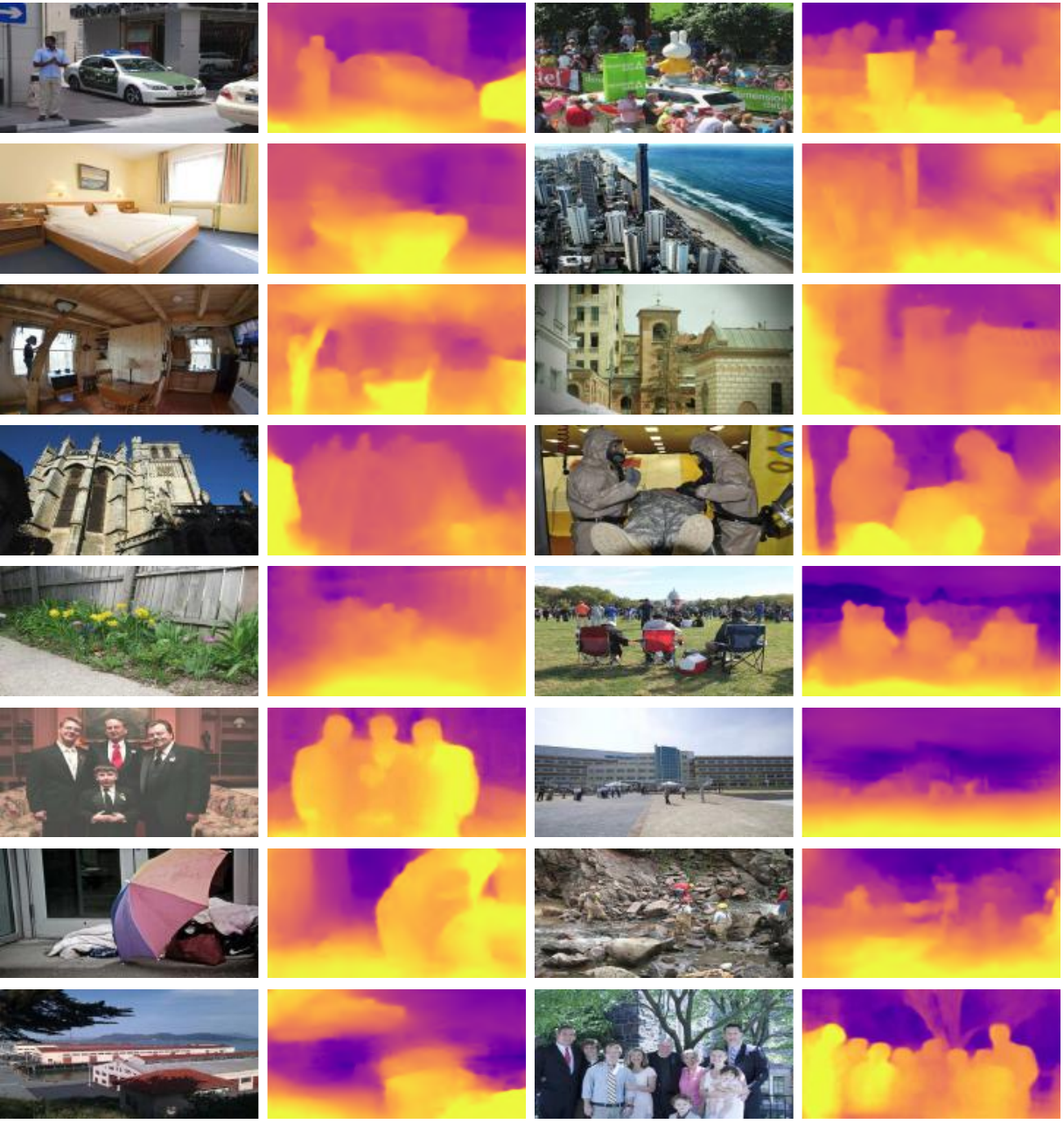}
\caption{Qualitative results on the DIW test set.}
\label{fig:example31}
\end{figure}

\begin{figure}[!ht]
\centering
\includegraphics[width=122mm]{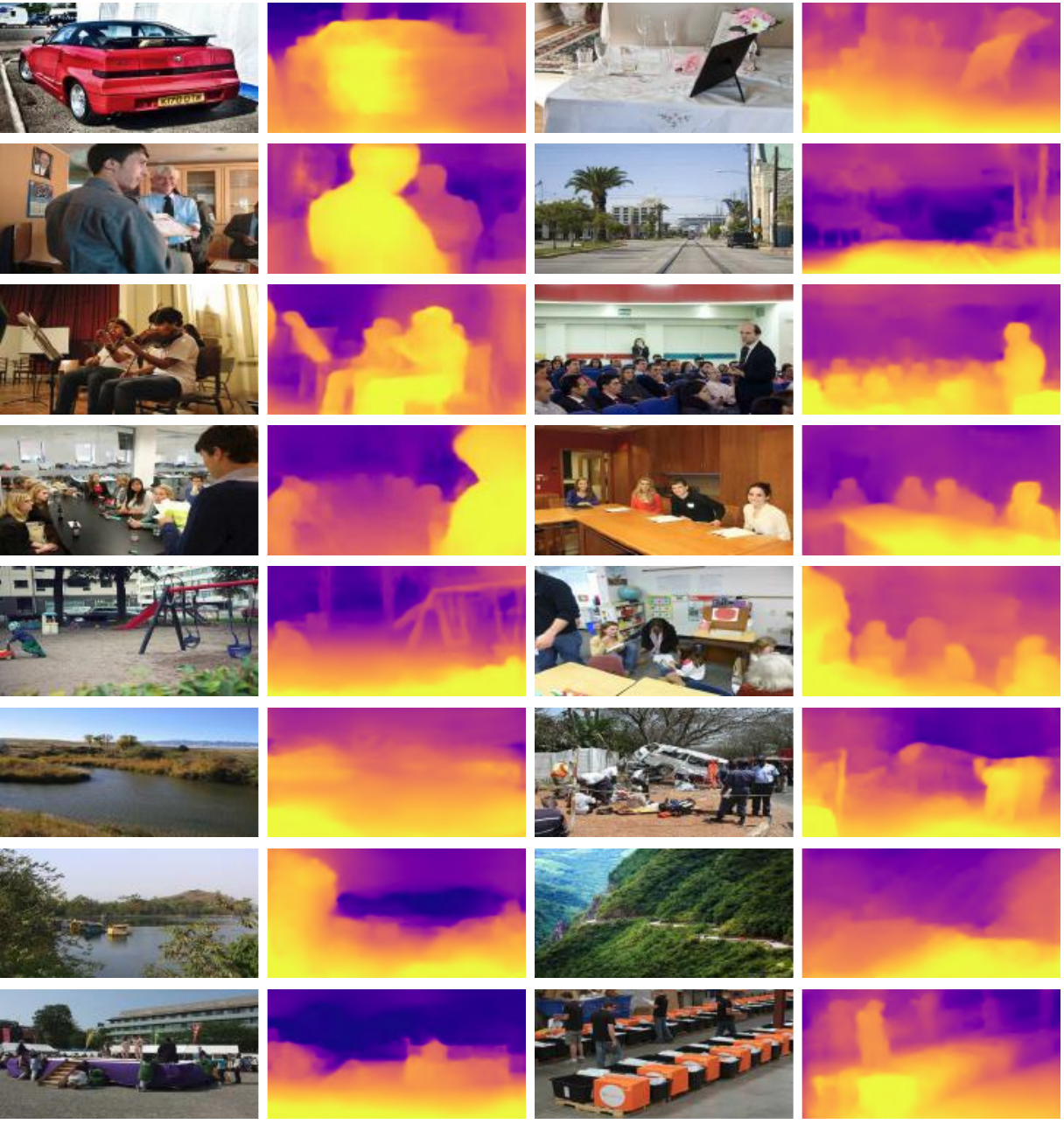}
\caption{Qualitative results on the DIW test set.}
\label{fig:example32}
\end{figure}

We provide additional qualitative results in order to showcase the generalization ability of the proposed model trained on MC+KITTI using our novel loss function. \\
\noindent\textbf{Supplementary video.} We generate depth predictions with our model and compare against current state-of-the-art Monodepth2 [19] on sample YouTube videos which consist of diverse scenes with dynamic objects and varying depth range. Results are presented in the supplementary video. Note: the video scenes are monocular and were not seen during training. These videos are recorded with standard handheld monocular cameras and do not have ground-truth depth estimates. Each frame was processed independently i.e. the temporal relation is not used. \\
\noindent\textbf{Diverse scene images.} We also show qualitative results on the test set of DIW [7] dataset in Fig. \ref{fig:example30}, \ref{fig:example31} and \ref{fig:example32}. These images constitute diversely rich content including indoor, natural and street scenes consisting of various objects taken from arbitrary camera angles with uncontrolled lighting conditions and scene appearance. Results demonstrate plausible depth estimation for general scenes with performance comparable to human or previous depth estimation using supervised learning [7].       

\end{document}